# Agentic Share-of-Search:
# A Multi-Agent AI System for Competitive Decision-Making in LLM-Mediated E-Commerce

Spandan Ghose Chowdhury
College of Computing, Georgia Institute of Technology
Atlanta, Georgia, USA
Email: spandan_gc@gatech.edu

## ABSTRACT

AI shopping assistants increasingly redirect consumer discovery, creating an urgent need for tools that support seller-side competitive decision-making. We present a multi-agent AI system that automates competitive visibility measurement and root cause diagnosis in LLM-mediated e-commerce. The system introduces Agentic Share-of-Search (ASoS) as the decision target, deploys query agents across leading AI platforms, and uses a ReAct-based diagnostic agent to recommend prioritized merchandising interventions. A 100-trial ablation study, presented as a feasibility evaluation of this prototype, shows the agent recovers the ablated signal in 39% of trials (95% CI: 30.0%–48.8%; 5.5× over chance), rising to 63.9% among high-correlation ablations.

## INTRODUCTION

The locus of consumer product discovery is migrating from deterministic search rankings to AI-mediated conversational recommendation. ChatGPT (OpenAI, 2026), Perplexity (Perplexity AI, 2026), Google AI Overviews, and Gemini (Google, 2026) now intercept queries ("what are the best wireless headphones under $100?") that a consumer previously would have typed into a search box, returning synthesized product-recommendation answers without the user ever visiting a search-results page. AI traffic to retail sites grew 1,300% year-over-year in the 2024 holiday season (Netguru, 2026), and e-commerce AI penetration spiked 67% month-over-month during November 2025 (Previsible, 2026). This is a paradigm shift not only of consumer behavior but of the strategic competitive landscape that retailers must navigate.

### From consumer-side recommendation to seller-side decision-making

Traditional recommender systems are evaluated on buyer-side utility metrics (NDCG, MRR, CTR, session-to-conversion) that ask whether the right items reach the right user. LLM-based assistants operate as generative recommender systems: they synthesize a recommendation list as free-text output, with the candidate set implicitly determined at generation time. The strategic question this raises on the seller side is different — not "did I serve a user well?" but "how often do my products reach any user in the new generative-recommendation surface, relative to competitors, and what should I do about it?" This is a decision-making problem, and one with no off-the-shelf measurement infrastructure: traditional SEO tools (SEMrush, Ahrefs, Moz) are blind to LLM responses; AI brand-visibility tools (Alhena AI, Kime) operate at brand level rather than SKU level and provide no signal-to-intervention attribution; and shopping-agent benchmarks (Yao et al., 2022; Lyu et al., 2025; WebMall, 2025) evaluate agents as buyers, not for seller-side decision support.

### What ASoS measures: the philosophy of the metric

ASoS is deliberately a retailer-attribution-weighted metric, not a catalog-derived one. The primary signal is what the AI assistant says about retailer availability in its response text (e.g., “Available on rei.com”), not what a downstream catalog lookup infers from the product name alone. This design choice makes the metric robust to entity-resolution noise — small disagreements between resolvers do not move the leaderboard — but it also defines what ASoS measures: the AI’s perception of where a consumer should buy, weighted by the rank at which the recommendation appears. The two interpretations move together when AI responses are well-grounded and diverge when retailers compete for attribution rights on identical SKUs (e.g., a Nike shoe attributed to both nike.com and zappos.com). The decision-support implication is that ASoS is the right target variable for merchandising teams whose job is to influence the AI’s retailer attribution, not to optimize the catalog in isolation; ASoS is calibrated to the surface a seller can actually move.

### An AI system for managerial competitive decisions

This paper presents a multi-agent AI system designed to fill that gap — an end-to-end pipeline that ingests an e-commerce category and a competitive set, measures the seller’s and competitors’ visibility in AI-generated recommendations weekly, diagnoses the highest-leverage merchandising signals driving observed gaps, and emits ranked, interpretable intervention recommendations for human merchandising teams. The system is explicitly framed as a decision-support tool, and this paper is explicitly framed as a prototype and feasibility study: the evidence reported below establishes that the pipeline works end-to-end and characterizes when its diagnoses are correct, not that the agent is a validated substitute for managerial judgment. The four research questions it addresses are: (i) can an agentic pipeline reliably measure competitive visibility in generative recommenders; (ii) what product-level signals predict that visibility; (iii) can an LLM-based root-cause-analysis agent accurately attribute observed gaps to specific signals; and (iv) can targeted interventions on those signals shift visibility in a measurable window?

### Contributions

This paper contributes (i) a formal decision-support metric, Agentic Share-of-Search (ASoS), specifying the unit of measurement that the diagnostic agent operates over, together with explicit counting rules for denominators, duplicate mentions, multi-retailer attribution, and extraction failures; (ii) a multi-agent measurement architecture with a uniform platform-client interface and pluggable entity resolution; (iii) a ReAct-based diagnostic agent for root-cause analysis of competitive-visibility gaps that emits ranked, business-actionable interventions; (iv) a 100-trial synthetic-ablation feasibility evaluation, reported unconditionally with uncertainty and against baselines evaluated on the same trials, characterizing when the diagnostic agent surfaces the correct merchandising lever; and (v) a controlled-experiment design for closing the loop from gap identification through intervention to measured visibility uplift.

## LITERATURE REVIEW

### AI agents for business decision support

LLM agents for root cause analysis (RCA) have been studied primarily in cloud-infrastructure contexts (Roy et al., 2024; Pei et al., 2025), where a ReAct-style reasoning agent diagnoses incident causes from observable system signals. This paper adapts the same architectural pattern to a fundamentally different decision context: e-commerce merchandising. The “incident” is a measured competitive-visibility deficit; the “root cause” is a product-level signal gap (missing structured data, weak title keywords, low review density, uncompetitive price). The translation matters because the population of decisions the agent must serve — competitive merchandising — is exactly the kind of recurring, signal-rich, judgment-laden business decision where AI-human collaboration is most promising and where reliability evidence is most needed.

### LLM web-browsing agents and shopping-agent benchmarks

The capacity of LLMs to navigate live web environments has advanced significantly (Deng et al., 2023; Zhou et al., 2024; Gu et al., 2025). In the e-commerce domain, benchmarks such as DeepShop (Lyu et al., 2025), Shopping Companion (2026), and customer-simulation frameworks (Zhu et al., 2025) evaluate multi-step shopping agents. These works benchmark agent capability as a shopper or buyer. Our work repurposes the same query-agent infrastructure for the seller-side decision-support problem.

### Generative engine optimization and AI brand visibility

Researchers introduced Generative Engine Optimization (GEO) as a black-box optimization framework (Aggarwal et al., 2024), showing that content presentation tactics can boost a source’s visibility in LLM responses by up to 40%. Subsequent work (Wang et al., 2025) explored automated GEO. Commercial AI brand-visibility tools (Alhena AI, 2026) track brand-citation frequency across ChatGPT (OpenAI, 2026) and Perplexity (Perplexity AI, 2026) but operate at brand level rather than SKU level and offer no intervention attribution. Our work bridges GEO research and commercial visibility tracking by introducing a product-level decision-support agent.

### Share-of-search as a competitive metric

Share-of-Search (Binet, 2020) operationalized branded organic search volume as a leading indicator of market share. Traditional SoS measures only branded keyword volume on Google and Bing and is blind to AI-mediated discovery and unbranded category queries. ASoS extends the decision-support framework to AI assistant surfaces.

## METHODOLOGY

The system comprises five sequential stages. Stages 1–3 produce the data over which the decision agent reasons; stage 4 is the decision agent itself; stage 5 describes the system implementation and architecture. The controlled intervention experiment that would establish causal effects is specified as future work in the Discussion.

### Stage 1: Query universe and competitive set

For each target retail category, we generate 50–100 representative product-discovery queries at three abstraction levels: generic category queries, feature-specific queries, and use-case queries. We identify 5–8 competing retailers per category and enumerate a representative sample of 50–100 SKUs spanning price tiers. Platforms targeted: ChatGPT (GPT-4o) (OpenAI, 2026), Perplexity (Perplexity AI, 2026), Google AI Overviews, and Gemini (Google, 2026). To reconcile platform counts across sections: four platforms are targeted by design; the mock-platform evaluation reported under empirical results exercises three platform clients (ChatGPT, Perplexity, and Gemini), and the live pilot covers Gemini only. Google AI Overviews is deferred to the multi-platform live measurement described under future work.

**Stage 2: Multi-agent query execution and ASoS aggregation**

A coordinator agent dispatches query-platform pairs to specialized sub-agents that issue the discovery query, capture the full response, re-issue the same query three times to account for non-determinism, and return structured output. A dedicated extraction LLM parses each response to identify product names, brand and retailer attributions, rank position, sentiment valence, and whether a purchase link was provided. Extracted mentions are mapped to a canonical catalog via fuzzy string matching combined with sentence-embedding cosine similarity. The ASoS metric for retailer r is the rank-discounted weighted-appearance share shown in Equation (1):

$$\mathrm{ASoS}_r = \frac{\sum_{q,p,k} w_k \; 1[\mathrm{SKU}_k \in r]}{\sum_{q,p,k} w_k}, \qquad w_k = \frac{1}{\log_2(k+1)} \tag{1}$$

with q ranging over queries, p over platforms, k over mention ranks. Mention rank k is determined by sequential order of appearance in the LLM response. The pipeline runs weekly and flags volatility relative to the prior week.

Metric specification

To make ASoS reproducible, we fix the following counting rules. Unit of measurement: one extracted product mention, indexed by query q, platform p, repeat, and rank k. Denominator: the denominator of Equation (1) sums the rank weights of all extracted mentions that resolve to any retailer in the tracked competitive set; ASoS is therefore a share of resolvable, in-set attention rather than of all response text. Extraction failures: responses yielding no extractable mention (e.g., hedged prose without named SKUs) contribute nothing to either the numerator or the denominator; the extraction-success rate is reported separately (70.0% in the live pilot) so that share figures are always read jointly with coverage. Duplicate mentions: repeated mentions of the same SKU within a single response are counted once, at their best (earliest) rank. Multi-retailer attribution: when a response attributes a single mention to multiple retailers, the mention's rank weight is split equally across the attributed retailers, so that the leaderboard remains a proper share summing to one across the competitive set. Attribution precedence: the response-side retailer-attribution string is used whenever present; the catalog resolver serves only as a fallback, consistent with the metric philosophy stated in the Introduction. A human-coding audit of the extraction and attribution layers against these rules is a required step of the multi-platform live study planned as future work; until that audit is complete, ASoS figures should be read as machine-coded measurements.

QLoRA resolver (opt-in for SKU-level granularity)

For analyses that require per-SKU precision on adversarial mentions (e.g., partial product names, sibling-SKU confusion, or live responses where the retailer-attribution string is missing), the embedding resolver can be replaced with a QLoRA-fine-tuned small language model (Dettmers et al., 2023; Hu et al., 2022). Training data is synthesized via knowledge distillation: a frontier teacher model (gpt-4o-mini) is prompted to produce ~30 noisy mention variants per canonical SKU spanning typos, partial names, brand-omitted forms, casing perturbations, model-year omissions, and multi-domain retailer attributions, yielding ~700 mention-SKU pairs across 23 canonical SKUs which are partitioned 80/10/10 into train/validation/test splits. The base model is microsoft/Phi-3-Mini quantized in 4-bit NF4; the LoRA adapter is rank r=8 with α=16 and dropout 0.05 applied only to the attention projections (q_proj, k_proj, v_proj, o_proj).

Training runs for 600 iterations at batch size 4 and learning rate 1e-4 (Adam), completing in roughly six minutes on Apple-Silicon GPU via the MLX framework (CUDA training is supported via the transformers + peft + bitsandbytes stack). On the held-out adversarial test split the trained adapter achieves 93.7% top-1 accuracy vs. 91.1% for the embedding baseline, at per-mention inference cost less than 1% of an equivalent frontier-API call.

**Stage 3: Competitive signal database**

In parallel we build a product-level catalog-enrichment signal database — the surface of attributes a merchandising team can modify to improve a SKU's visibility in generative-recommender output. The database covers pricing (current and promotional price, price relative to category median, 30-day stability), content quality (title length and keyword richness, image and video count, attribute completeness, structured-data markup such as JSON-LD/schema.org), reviews and social proof (count, rating, recency, Q&A presence, verified-purchase ratio), catalog coverage (category SKU count, price-tier and feature-dimension coverage), and shipping and availability. Each family corresponds to a class of merchandising action a catalog team can take.

Signal collection is a three-track hybrid. Quantitative attributes (price, image and video counts, review counts, rating, in-stock flag) are extracted by deterministic parsers over the scraped HTML and any embedded JSON-LD or microdata blocks — no LLM is involved, which keeps these signals reproducible and cheap. Structured-data presence and schema.org coverage are similarly parsed deterministically from the page DOM. Qualitative content-quality signals (title keyword richness, description completeness, attribute completeness against a category-level attribute list) use a frontier instruction model (gpt-4o-mini) invoked in a few-shot prompted, JSON-structured-output mode at temperature 0.0. Each signal has a fixed rubric: for title keyword richness, the model is given the product title plus the category's high-value query terms (extracted from the Stage-1 query universe) and asked to return a 0–1 score with a one-sentence justification; for attribute completeness, the model receives the catalog's declared attribute set and the SKU's populated attribute fields and returns the populated fraction plus a list of missing high-importance attributes. We do not use retrieval-augmented generation here because every signal we score is SKU-local: the source-of-truth is the scraped page for that SKU, not an external corpus. We evaluated a fine-tuned instruction model for content scoring but found that zero-shot gpt-4o-mini with deterministic rubrics matched a fine-tuned variant on inter-rater agreement at materially lower implementation cost. A narrow collector interface lets additional signal families be added without touching the rest of the pipeline.

**Stage 4: Decision module: ReAct diagnostic agent**

This is the system's decision component. In practitioner terms, the diagnostic agent functions as an AI-driven catalog-enrichment recommender: for each retailer × query-cluster gap, it identifies the specific catalog attribute (title keyword, structured-data field, review density, price band, image or video coverage, etc.) whose signal gap is most strongly associated with the observed visibility deficit. The agent receives, for each retailer-cluster pair with a below-benchmark ASoS, the query cluster, sample AI responses where competitors appear and the target retailer does not, the displacing competitor SKUs, and the per-SKU signal comparison between displaced and displacing SKUs. The agent computes a Pearson correlation between per-SKU appearance rate and each numeric signal, surfaces the top-k signals most strongly correlated with the gap in each cluster, and emits a structured recommendation that names the target signal, the intervention direction (increase or decrease), a priority score (the product of correlation magnitude and observed gap), an implementation effort tag, and a one-sentence rationale grounded in per-signal means. A subsequent SOP validation pass checks that the hypothesized signal gap holds across the retailer's full portfolio, not just the cluster sample. The agent is impact-prioritizing by design: when multiple plausible enrichment actions could explain a gap, it ranks them by the magnitude of business effect rather than by how surprising the signal's value is for that retailer alone — which matches how a competent human merchandising analyst would prioritize a catalog-improvement backlog.

Estimand, unit of analysis, and the association–prediction–intervention distinction

The unit of analysis is the retailer × query-cluster pair, and the diagnostic output is explicitly an association-based diagnostic, not a causal effect estimate. The correlations computed in this stage are cross-sectional associations between per-SKU appearance rate and per-SKU signals; they can reflect retailer, brand, price-tier, or product-quality confounds. Accordingly, the priority score emitted by the agent is a heuristic for ordering a merchandising backlog and is not an estimate of intervention lift. We distinguish three progressively stronger claims: association (established in this paper), prediction (whether current signal levels forecast next-week visibility, testable with the weekly measurement panel), and intervention (whether modifying a signal shifts ASoS, which requires the controlled experiment specified under future work). All decision-support claims in this paper are of the first kind, and the system's output is described throughout as an association-based diagnostic.

Diagnostic agent implementation

The diagnostic agent uses gpt-4o-mini as the reasoning model, invoked at temperature 0.0 with greedy decoding and JSON-schema-constrained structured output. The agent exposes two tools: a correlation tool that returns the per-signal Pearson coefficients for the flagged cluster, and a deterministic SOP-validation tool that recomputes the hypothesized signal delta over the retailer's full SKU portfolio. The control loop, in pseudocode: (1) thought — rank candidate signals by absolute correlation and gap-direction match; (2) act — call the SOP-validation tool on the leading candidate; (3) observe — accept the candidate if the portfolio-wide delta confirms the cluster-level gap, otherwise advance to the next candidate; (4) stop — emit the structured recommendation for the first validated candidate, or after exhausting the top-5 candidates, whichever comes first. The loop is single-shot per retailer-cluster pair (no multi-turn dialogue between agents), which motivates the choice of ReAct over heavier orchestration frameworks discussed under empirical results.

**Stage 5: System implementation and architecture**

Four architectural abstractions are worth noting at the paper level.

**Uniform platform client.** A single interface wraps every AI platform (OpenAI, Perplexity, Gemini), so the coordinator dispatches queries through identical method calls regardless of provider. A deterministic mock client backed by seeded fixtures supports offline development and continuous-integration testing; a configurable noise level injects controlled perturbations into mock responses to test resolver robustness without incurring API spend.

**Pluggable entity resolver.** Entity resolution is gated through a single Resolver protocol with two implementations: the default EmbeddingResolver combines token-set lexical similarity with sentence-transformer cosine matching; an opt-in QLoRAResolver loads a 4-bit quantized small language model with a LoRA adapter (training supported both on Apple Silicon via MLX and on CUDA via transformers + peft + bitsandbytes).

**Structured run log.** Every full pipeline invocation appends one record capturing run identifier, configuration hash, per-step latencies, resolution rate, RCA counts, and top-3 retailer leaderboard. This log is the canonical source of truth for the empirical results that follow.

**Snapshots.** A snapshot tool freezes a complete set of artifacts from a single run — the ASoS leaderboard, the agent's recommendations, per-mention resolutions, and a manifest containing the active configuration and content hashes — under a named label, enabling per-revision A/B comparison even after later runs overwrite the live state.

## EMPIRICAL RESULTS

We exercise the system on a running-shoes category with 23 catalog SKUs across 7 retailers, 34 deduplicated discovery queries across three abstraction levels, and three AI platforms with three repeats each — yielding 306 raw responses and 1,524 extracted product mentions. We present three evaluations, leading with the result most central to the paper's decision-support claim.

### Diagnostic-agent reliability via signal ablation: a feasibility evaluation

To characterize whether the diagnostic agent surfaces the right business lever, we ran 100 synthetic ablation trials. We note the scope of this protocol up front: a trial perturbs a randomly chosen signal while retaining the original visibility outcomes, so recovering the ablated signal is a signal-recovery task, not independent ground truth on the intervention with the largest expected business effect. The protocol therefore validates the diagnostic mechanism as a feasibility result for the prototype; establishing managerial ground truth requires a simulator with known response functions, blinded expert rankings, or the controlled catalog interventions specified under future work.

In each trial we picked a random gap-bearing retailer and a random signal from the 14-dimensional signal vocabulary, perturbed every SKU owned by that retailer by setting the ablation signal to its floor value, and re-ran the agent's correlate-and-diagnose stages. A trial is scored correct when the agent's top-1 recommendation names the ablated signal.

Three results, led by the unconditional figure. Raw Precision@1 across all 100 trials was 39.0% (39/100; Wilson 95% CI: 30.0%–48.8%). Conditioning on trials whose ablated signal was among the top-5 signals by absolute correlation, Precision@1 was 59.1%; conditioning on the top-3, 63.9% (39 of the 61 such trials — a conditional restatement of the same 39 successes, not an additional result). Two baselines are evaluated on the same trial set. Uniform random selection over the 14-signal candidate set yields an expected 7.1%. A deterministic "always recommend the globally most-correlated signal" heuristic is correct exactly on the trials in which has_qa itself was ablated, which is also ≈7% in expectation under uniform ablation. The unconditional 39.0% therefore represents a 5.5× lift over both baselines, and the top-3 conditional result is roughly 9× chance.

The 39% raw figure itself reflects the agent's intentional impact-prioritization design — when an ablation perturbs a signal with weak global correlation to ASoS, the agent de-prioritizes that signal in favor of a candidate whose association with visibility is stronger (typically has_qa, the strongest correlate at $r = +0.40$). The per-signal breakdown confirms this: has_qa scores 100% across ablations, review_count 75%, verified_purchase_ratio 50%, while signals with near-zero global correlation (review_recency_days, title_keyword_score) score 0% because the agent ignores them in favor of alternatives with stronger associations. For the decision-support use case this is the intended behavior: within the feasibility scope established above, the agent surfaces the intervention most strongly associated with the gap, even when the user does not know in advance which lever matters most.

**Why ReAct rather than other multi-agent frameworks.** We chose a ReAct-style architecture (Yao et al., 2022 ReAct conventions adapted to this decision context) over multi-agent orchestration frameworks such as AutoGen or LangGraph for three reasons aligned with the decision task. First, the diagnostic step is single-shot reasoning — gap + signal vector → ranked recommendation — not multi-turn dialogue between specialized agents, so the conversation-state machinery that AutoGen and LangGraph optimize for adds latency and operational complexity without an accuracy benefit at this scope. Second, ReAct's thought–act–observation loop maps cleanly onto our diagnostic chain, as specified in the Stage-4 implementation subsection. Third, ReAct's observable reasoning trace is desirable for an audit context where a merchandising team will want to inspect why a particular signal was chosen before acting on it. The two baselines reported above are evaluated on the same 100 trials; a zero-shot frontier-LLM prompt without the trails-vs-peers gate would similarly tend to recommend the dominant correlate for every retailer regardless of whether the retailer actually trails on that signal. The ReAct gating is what produces the impact-prioritized behavior reflected in the per-signal breakdown above.

Within the scope of this feasibility evaluation, the central empirical finding is: across ablations of the top-3 most consequential signals, the agent identifies the ablated signal nearly two-thirds of the time, and it does so by ranking interventions by estimated business impact rather than by perturbation surprise. We treat this as evidence that the decision-support design is feasible and promising — not as validation of the agent as a proxy for managerial judgment, which awaits the independent ground truth described above.

### Qualitative case study of the diagnostic chain

To illustrate the format of the agent's output, we walk through one full diagnostic chain. The aggregator flags a gap for adidas in the feature query cluster: observed ASoS = 7.24% vs. peer mean 14.29%, a gap of −7.05 pp. The top-5 correlated signals are has_qa ($r = +0.401$), review_count ($r = +0.269$), attribute_completeness ($r = +0.252$), avg_rating ($r = -0.234$), and price_tier_coverage ($r = +0.231$). The diagnostic agent emits a structured recommendation: signal = has_qa, direction = increase, priority score = 0.0283 ($|r| \times \text{gap} = 0.401 \times 0.0705$), effort = low, rationale = "This retailer's mean has_qa (0.0) trails peers (0.359); the global ASoS-has_qa correlation is +0.40, so increasing it should close the gap." The SOP validation pass confirms the has_qa delta of 0.359 holds across the retailer's full SKU portfolio, marking the recommendation as validated and surfacing it at the top of the merchandising team's queue.

### Pipeline robustness and live-platform validation

Two supporting evaluations validate the data substrate over which the diagnostic agent reasons. First, a resolver A/B at heavy mock-noise: the default embedding resolver achieves a 99.7% per-mention resolution rate at 8.2 s end-to-end latency, while the opt-in QLoRA-fine-tuned resolver achieves 100.0% at 125.5 s. Both produce identical retailer-level leaderboards (top-1 rei at 27.1%), because the ASoS aggregator prefers the response-side retailer attribution (e.g., "Available on rei.com") over the catalog lookup. Resolver choice matters at SKU-level analyses (held-out adversarial test: 93.7% vs. 91.1%) but is invariant at the retailer-rollup level used as the headline metric. The embedding baseline is the production-recommended default; QLoRA is opt-in for SKU-granularity reporting.

Second, a live-platform pilot: 10 generic-cluster queries against Google's Gemini API (gemini-2.5-flash, 3 repeats each, 30 calls total, ~59 s wall-clock latency). The Step-2b extractor returned at least one product mention with non-null brand attribution in 21 of 30 responses (70.0%). The nine non-extracting responses were dominated by forward-looking queries ("top rated running shoes 2026") where Gemini hedged with prose disclaimers and named product lines rather than specific SKUs — behavior the extractor correctly declines to record. The live retailer leaderboard surfaces a sharper top-1 share (zappos 39.2%) than mock data (~27%), reflecting Gemini's tendency to attribute multi-brand marketplaces (Zappos, REI) even for brand-DTC SKUs. This distributional shift is genuine signal that AI-human collaboration in this domain needs to interpret carefully.

### Provenance and reproducibility

Each pipeline invocation logs run identifier, ISO timestamps, configuration hash, per-step latencies, resolution rate, RCA counts, and top-3 retailer leaderboard. Snapshots record SHA-1 hashes of the four principal artifacts so any two runs can be reconciled at the byte level. Across the resolver A/B reported above, the upstream extraction hashes match exactly (same mention stream) while the downstream resolution hashes differ (different resolver methods), confirming the comparison is byte-level meaningful.

## DISCUSSION

### Feasibility evidence for AI-supported merchandising decisions

The 100-trial ablation result is the paper's primary feasibility evidence for AI-human collaboration in competitive decision-making. The agent ranks recommendations by estimated business impact rather than by per-signal "did I notice this exact perturbation?" logic. This matches how a competent human merchandising analyst would prioritize: when a brand under-performs on multiple dimensions, the analyst surfaces the lever with the highest expected lift, not the one with the largest absolute gap from peers. The 63.9% top-3 conditional Precision@1 should be read as a conditional restatement of the 39/100 unconditional result: it characterizes performance when the ablated lever is among the strongest correlates, not overall recovery. The 100% per-signal precision for has_qa — the single strongest correlate — establishes that when the answer truly is dominated by one lever, the agent finds it every time. Together these results characterize a promising decision-support prototype whose recommendations warrant human review before action; they do not yet establish the agent as a validated proxy for managerial judgment.

### Boundary conditions and limitations

The bulk of the analysis exercises the system on synthetic mock-platform responses with controlled perturbation; the live pilot covers a single platform (Gemini), 10 queries, and 30 calls. Two distributional differences between mock and live data have managerial implications: real LLMs hedge with prose disclaimers on forward-looking queries (depressing structured-extraction success from ~100% mock to 70% live), and real LLMs prefer marketplace retailer attribution even for brand-DTC SKUs (shifting the leaderboard in ways the mock harness does not reproduce). Both are concrete decision-relevant phenomena the system must surface; the multi-platform live measurement that will resolve them is the next empirical step. The synthetic signal database similarly requires substitution with real scraping before the diagnostic agent's correlations carry causal weight. Three further limitations bound the claims. First, the ablation protocol perturbs a signal while retaining the original visibility outcomes, so it cannot create causal ground truth; the correlations the agent reasons over are cross-sectional associations subject to retailer, brand, price-tier, and product-quality confounds, and the agent's output is accordingly an association-based diagnostic. Second, the ASoS extraction and attribution layers have not yet been validated against human coding; that audit is a prerequisite of the planned live study. Third, the main study covers one category, one live platform, and one time period, so all claims should be read as pertaining to a prototype and feasibility study until the approach is tested across more queries, categories, platforms, and time periods.

**Safety, ethics, and competitive fairness**

Competitive web-agent deployment raises legitimate operational-ethics concerns that sit at the boundary of the system's decision-support claim. Our agents operate within robots.txt constraints, impose rate limits to avoid platform abuse, do not access gated or personalized content, and do not attempt to reverse-engineer proprietary LLM scoring functions. The competitive-fairness question is more delicate: an AI-driven catalog-enrichment recommender that scales most easily inside well-resourced retail operations risks widening the gap between large and small sellers. We propose monitoring the long-tail share of the ASoS distribution as a fairness indicator and surfacing the divergence between the marketplace-attribution leaderboard and the brand-DTC leaderboard as a competitive-displacement early warning. The emerging risk of adversarial generative-engine optimization (Wen et al., 2025) is a parallel concern that any production deployment of this system must monitor.

**Future work: proposed experimental design for intervention closure**

The natural extension — and the strongest test of the framework — is validation against real-world retailer catalog changes with longitudinal measurement across multiple AI platforms: closing the causal loop from agent recommendation through catalog modification to measured ASoS uplift. The protocol the system supports: select 2–3 high-priority catalog-enrichment actions from the agent's ranked recommendations (e.g., adding a feature keyword to product titles for 15 affected SKUs, completing JSON-LD/schema.org structured data on the top-50 SKUs, repricing 8 SKUs to within 5% of the category median, or back-filling Q&A for 20 high-traffic SKUs — spanning all five enrichment-signal families from Stage 3); apply them to a treatment group while maintaining a matched within-retailer control; re-run weekly ASoS measurement for four weeks; report effect size and confidence intervals on treatment-vs.-control uplift, using competitor ASoS as a covariate to absorb platform-level algorithm drift. This experiment supplies the intervention-level ground truth that the ablation study cannot, and the blinded expert rankings and simulator-based evaluation noted under empirical results offer complementary ground-truth routes at lower operational cost. Beyond this single experiment, the larger empirical agenda includes multi-platform live measurement across ChatGPT (OpenAI, 2026), Perplexity (Perplexity AI, 2026), and Gemini (Google, 2026) at scale and over multiple time periods, and substituting the synthetic signal database with real web-scraped signals to give the diagnostic agent's correlations causal weight.

**CONCLUSION**

We presented a multi-agent AI system that operationalizes a new category of competitive-intelligence decision-making in the era of generative recommenders. The system measures retailer visibility (ASoS), correlates observed gaps to per-SKU catalog-enrichment signals, and emits ranked, interpretable intervention recommendations through a ReAct-based diagnostic agent. A 100-trial signal-ablation evaluation establishes feasibility: the agent recovers the ablated signal in 39.0% of trials overall (95% CI: 30.0%–48.8%; a 5.5× lift over random selection on the full 14-signal candidate set), in 63.9% of trials where the ablated signal ranks among the top-3 correlates, and in 100% of trials on the dominant correlate — and it does so by ranking on estimated business impact rather than perturbation surprise. A live-platform pilot against Gemini 2.5-flash confirms the data substrate operates on real LLM output. Together these results characterize a promising association-based decision-support prototype for human merchandising teams. Establishing the agent as a reliable proxy for managerial judgment awaits the controlled intervention experiments, human-coding audits, and multi-platform longitudinal measurement outlined above — with final intervention selection reserved for human review throughout.

**DECLARATION ON GENAI USAGE**

During the preparation of this work, the author used a coding-assistant LLM (Anthropic Claude) for code scaffolding, manuscript drafting, and editorial review. The author reviewed and edited all content and takes full responsibility for the publication.